\documentclass[sigconf,nonacm]{acmart}

\usepackage{graphicx}
\graphicspath{{../figures/}}
\usepackage{booktabs}
\usepackage{amsmath}
\usepackage{algorithm}
\usepackage{algpseudocode}

\renewcommand\footnotetextcopyrightpermission[1]{}
\begin{document}

\title{Pathology Transport: Optimal-Transport Explanations for Clinical Data,\\
and When Their Heatmaps (Fail to) Localize Disease}

\author{Lalit Kumar}
\affiliation{%
  \institution{Department of Computer Science\\
  The University of Texas at Austin\\lalit.kumar@utexas.edu}
  \country{}}
\email{}

\begin{abstract}
Generative models promise a route to explainable clinical AI: rather than probe a classifier,
model the distributions of \emph{healthy} and \emph{diseased} patients and read explanations off
the geometry between them. We build such a system---an optimal-transport \emph{rectified flow}
trained between two clinical distributions---and use it to ask a pointed question the field too
rarely tests: do the resulting explanation \emph{heatmaps actually localize disease}? On tabular
tumour biomarkers (Breast Cancer Wisconsin) a single flow yields per-patient counterfactuals, an
unsupervised malignancy score (AUROC~0.91; $0.93\pm0.01$ across five seeds), and a label-free
attribution that agrees with a supervised classifier ($r{\approx}0.5$)---a compact, honest
interpretability engine, though it never out-predicts logistic regression. Moving to chest
X-rays, we show the transport heatmap is a \emph{population-level} signal, not a localiser; a
reconstruction-based, identity-preserving variant \emph{does} localize \emph{synthetic} lesions
(pointing game $0.52$), yet on \emph{real} RSNA radiologist boxes it collapses to chance while
only supervised Grad-CAM stays above it. The central result is a \emph{synthetic-to-real gap}:
label-free heatmaps that look compelling on planted lesions are not evidence of real
localisation. We contribute a reusable optimal-transport recipe for generative explanations and
a controlled benchmark for stress-testing whether they localize.
\end{abstract}


\maketitle

\section{Introduction}
Clinical adoption of predictive models is limited less by accuracy than by
\emph{trust and actionability}. A model that outputs ``87\% malignant'' gives a clinician
a number but not a rationale: which measurements drove the decision, and what would have
to be different for the verdict to change? Explainable-AI methods such as SHAP and
saliency answer the first question with feature-importance weights, but they do not
produce a concrete, on-distribution example of the counterfactual patient. Counterfactual
explanations---``the smallest change to the inputs that flips the prediction''---answer
exactly this question and are increasingly seen as the form of explanation clinicians and
regulators actually want~\cite{wachter2017}.

The dominant way to obtain counterfactuals is to perturb the input against a fixed
classifier. We ask a different, higher-risk question: \emph{what if we never train a
classifier at all, and instead learn the geometry that separates health from disease
directly?} Concretely, we treat the benign and malignant patient populations as two
probability distributions in biomarker space and learn the \emph{optimal-transport map}
that carries one onto the other. Recent advances in generative modelling make this
practical: \emph{rectified flow}~\cite{liu2023} and \emph{flow
matching}~\cite{lipman2023} learn a velocity field whose ordinary differential equation
transports one distribution into another by simple regression, and mini-batch
optimal-transport coupling~\cite{tong2024} makes those trajectories nearly straight and
stable.

Our thesis is that a \emph{single} such transport model is a surprisingly complete
clinical explanation engine. The map itself is a per-patient counterfactual generator;
the length a patient must travel is an unsupervised risk score; and the average
displacement is a global attribution. We test this on the Breast Cancer Wisconsin
(Diagnostic) dataset~\cite{street1993}---569 biopsies, 30 nuclear biomarkers---chosen
because it is real clinical data, is fully offline, and has a known ground-truth
biomarker signature against which our label-free attribution can be validated.

This is a high-risk design and we treat it as such. Learned distribution transport can
overshoot and fabricate off-manifold ``patients''; an unsupervised score has no guarantee
of matching a supervised classifier; and near-linearly-separable data is a regime where
simple linear methods are notoriously hard to beat. We report where the method succeeds
and, just as clearly, where it does not. The imaging half then presses a sharper question---do
these generative heatmaps genuinely \emph{localize} disease?---which we answer with a controlled
synthetic benchmark and a real RSNA annotation test.

\paragraph{Contributions.}
\begin{itemize}
\item We reframe tumour diagnosis as optimal transport between clinical distributions and
implement it with an OT-coupled rectified flow (Section~\ref{sec:method}).
\item From one model we derive three interpretability artefacts---counterfactuals, an
unsupervised risk score, and population attribution---and evaluate each quantitatively
(Section~\ref{sec:results}).
\item We give an honest account of failure modes: no accuracy gain over logistic
regression, non-sparse edits, and a score whose discrimination partly reflects
off-manifold drift.
\item We show the recipe is modality-agnostic and, via a controlled synthetic-lesion
benchmark and a \emph{real} RSNA bounding-box test, deliver a cautionary finding: a
label-free heatmap that localises synthetic lesions fails to transfer to real pathology.
\end{itemize}

\section{Related Work}
\paragraph{Counterfactual explanations.}
Wachter et al.~\cite{wachter2017} formalised counterfactual explanations as an
optimisation that finds the nearest input flipping a model's decision, and argued they
satisfy legal ``right to explanation'' requirements without exposing model internals.
Subsequent work adds validity, sparsity and plausibility constraints~\cite{guidotti2024}. Our approach
differs in that the counterfactual is produced by a \emph{generative transport map}
rather than by gradient descent against a classifier, so it is defined even when no
classifier exists and is naturally biased toward the real data manifold.

\paragraph{Explainable AI in healthcare.}
Feature-attribution methods such as SHAP~\cite{lundberg2017} have become the default lens
for interpreting clinical risk models, assigning each input a contribution to the
prediction. They are, however, tied to a trained predictor and explain \emph{a model's
decision} rather than \emph{the disease}: they cannot synthesise the patient who would
receive a different verdict, nor localise pathology in an image without a supervised
detector. Our transport-based view is complementary---it explains the \emph{data geometry}
separating health from disease, and a single object yields counterfactuals, a score, and
spatial attribution at once.

\paragraph{Generative and diffusion counterfactuals.}
A growing line of work generates counterfactuals with deep generative models, especially
in medical imaging, e.g.\ diffusion-based counterfactuals that morph a diseased scan into
its healthy version to localise pathology~\cite{jeanneret2022}. These methods still
condition on an external classifier for guidance. We instead let the transport between
two \emph{unconditional} class distributions define the edit, which is closer in spirit
to population-dynamics models that use optimal transport to interpolate biological state,
such as TrajectoryNet for single-cell trajectories~\cite{tong2020}.

\paragraph{Saliency, localisation, and faithfulness.}
For images we compare against \emph{Grad-CAM}~\cite{selvaraju2017}, the standard supervised
saliency method, and quantify map quality with deletion/insertion faithfulness~\cite{petsiuk2018}
and a pointing-game/IoU protocol against ground-truth boxes. Our label-free localiser is a
\emph{reconstruction-based} anomaly detector in the spirit of f-AnoGAN~\cite{schlegl2019} and
autoencoder anomaly segmentation~\cite{baur2021}---a model of healthy anatomy whose
reconstruction error flags the abnormal, an approach whose strongest variants now use
diffusion restoration~\cite{wyatt2022}---which we evaluate on real annotated pneumonia from
the RSNA Pneumonia Detection Challenge~\cite{rsna2019}.

\paragraph{Flow matching and optimal transport.}
Rectified flow~\cite{liu2023} and flow matching~\cite{lipman2023} learn a velocity field
by regressing onto the direction of straight-line interpolations between paired samples,
avoiding the simulation of stochastic diffusion. Tong et al.~\cite{tong2024} show that
choosing the pairing via mini-batch optimal transport straightens trajectories and
improves sample quality; this coupling is central to our method's stability. To our
knowledge, using an OT-coupled rectified flow \emph{between two clinical class
distributions} to jointly yield counterfactuals, a risk score, and attribution has not
been reported.

\section{Background}
\label{sec:background}
\paragraph{Optimal transport.}
Given a source distribution $\mu$ and a target $\nu$, optimal transport seeks the map (or
plan) that morphs one into the other at minimum total cost. Under the squared-Euclidean
cost the Monge problem is $\min_{T:\,T_\#\mu=\nu}\int\|x-T(x)\|^2\,d\mu(x)$, and its
Kantorovich relaxation optimises over couplings $\pi\in\Pi(\mu,\nu)$, minimising
$\int\|x_0-x_1\|^2\,d\pi(x_0,x_1)$. Intuitively, OT pairs each source point with the target
point it can reach most cheaply, so the ``movement'' it prescribes is the most
economical---and therefore most interpretable---transformation of one population into the
other. We never form the full continuous plan; we approximate it per mini-batch with a
discrete assignment (Section~\ref{sec:method}).

\paragraph{Rectified flow.}
A rectified flow~\cite{liu2023} represents transport as an ordinary differential equation
$\dot x = v_\theta(x,t)$ that carries samples of $\mu$ at $t{=}0$ to samples of $\nu$ at
$t{=}1$. Rather than simulating a stochastic process, it \emph{regresses} the velocity
onto the direction of a straight line between paired endpoints---the flow-matching
objective~\cite{lipman2023}. When the endpoints are paired by optimal
transport~\cite{tong2024}, the target velocities have low variance and the learned
trajectories are nearly straight, so a coarse Euler integrator suffices and the net
displacement $x-F(x)$ is a faithful estimate of the transport. That displacement is the
quantity we mine for every downstream artefact.

\section{Methodology}
\label{sec:method}

\begin{figure*}[t]
\centering
\includegraphics[width=\textwidth]{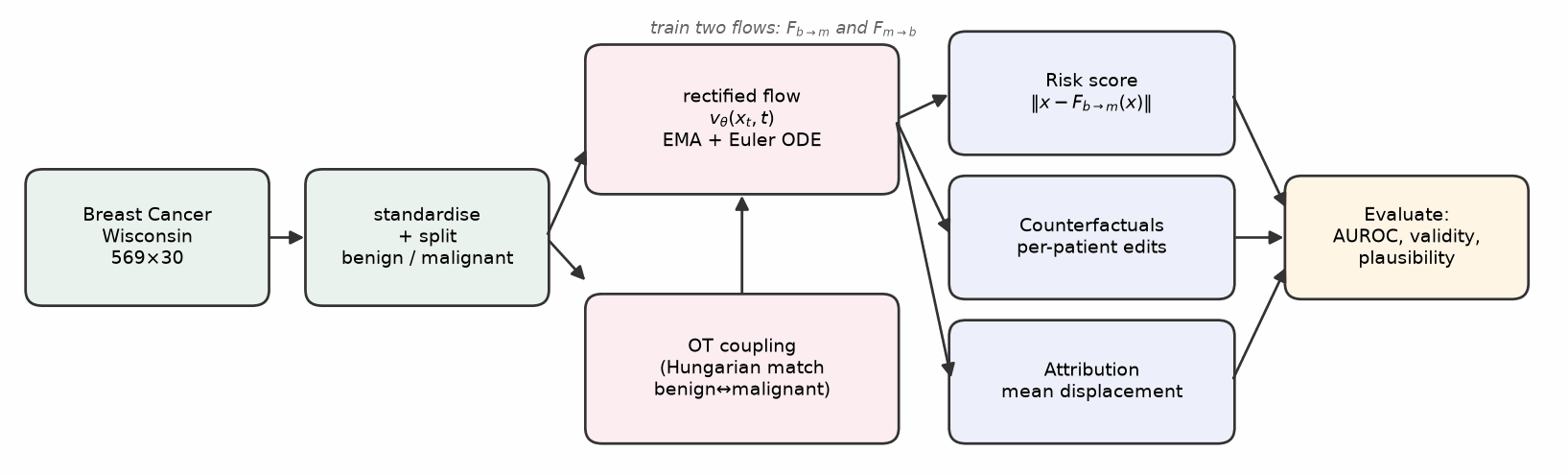}
\caption{Workflow. Standardised biomarkers are split into benign (source) and malignant
(target) sets. Within each mini-batch a Hungarian optimal-transport assignment pairs
source and target points; a time-conditioned velocity field $v_\theta(x_t,t)$ is
regressed onto the pairing direction (rectified flow) and integrated with an Euler ODE
under EMA weights. Two flows, $F_{b\to m}$ and $F_{m\to b}$, yield a risk score,
per-patient counterfactuals, and population attribution, each evaluated against
supervised or naive baselines.}
\label{fig:workflow}
\end{figure*}

\paragraph{Data.}
The Breast Cancer Wisconsin (Diagnostic) dataset contains 569 fine-needle-aspiration
biopsies, each described by 30 real-valued nuclear biomarkers (the mean, standard error
and ``worst'' of ten cell-nucleus measurements). We label malignant as the positive
class, standardise features using statistics from the training split only, and hold out
30\% for evaluation. Working in $z$-score space makes a unit of transport displacement
comparable across biomarkers.

\paragraph{Rectified flow.}
We learn a velocity field $v_\theta(x,t):\mathbb{R}^{30}\times[0,1]\to\mathbb{R}^{30}$
whose ODE $\dot x = v_\theta(x,t)$ transports the benign distribution at $t{=}0$ into the
malignant distribution at $t{=}1$. Following rectified flow, we train on straight-line
interpolations between paired endpoints:
\begin{equation}
x_t=(1-t)\,x_0+t\,x_1,\quad
\mathcal{L}(\theta)=\mathbb{E}_{x_0,x_1,t}\big\|v_\theta(x_t,t)-(x_1-x_0)\big\|^2 .
\end{equation}
$v_\theta$ is a four-layer SiLU multilayer perceptron with a sinusoidal time embedding.

\paragraph{Optimal-transport coupling.}
The choice of pairing $(x_0,x_1)$ is decisive. Independent random pairing yields
extremely high-variance regression targets, so the integrated ODE overshoots and lands
off-manifold. We instead use mini-batch optimal-transport coupling: within each batch we
solve the assignment $\min_\pi \sum_i \|x_0^{(i)}-x_1^{(\pi(i))}\|^2$ with the Hungarian
algorithm and pair points by their transport-optimal match~\cite{tong2024}. In our
experiments this alone cut the counterfactual edit size by $2.6\times$ and turned an
incoherent attribution ($r{=}-0.03$) into a clinically sensible one ($r{=}0.49$). We keep
an exponential-moving-average copy of the weights for stable integration and train both
directions, $F_{b\to m}$ and $F_{m\to b}$, with a 100-step Euler integrator
(Figure~\ref{fig:workflow}).

\begin{algorithm}[t]
\caption{OT-coupled rectified-flow training}
\label{alg:train}
\begin{algorithmic}[1]
\Require source set $A$, target set $B$, steps $N$, batch size $b$
\State initialise $v_\theta$; EMA copy $\bar\theta \gets \theta$
\For{$i=1$ to $N$}
  \State sample $\{x_0^k\}\sim A$, $\{x_1^k\}\sim B$, $k=1..b$
  \State $C_{kl} \gets \|x_0^k - x_1^l\|^2$ \Comment{cost matrix}
  \State $\pi \gets \textsc{Hungarian}(C)$ \Comment{mini-batch OT pairing}
  \State $x_1 \gets x_1[\pi]$
  \State $t \sim \mathcal{U}(0,1)$;\; $x_t \gets (1-t)x_0 + t\,x_1$
  \State $\mathcal{L} \gets \|v_\theta(x_t,t) - (x_1-x_0)\|^2$
  \State $\theta \gets \text{Adam}(\nabla_\theta \mathcal{L})$;\; $\bar\theta \gets 0.999\,\bar\theta + 0.001\,\theta$
\EndFor
\State \Return EMA model $F = \bar\theta$
\end{algorithmic}
\end{algorithm}

\paragraph{Three artefacts.}
\emph{(1) Risk score.} We score a patient by the magnitude of the benign$\to$malignant
transport applied to it, $s(x)=\|x-F_{b\to m}(x)\|$. A genuinely malignant patient lies
off the benign source support, so the ODE drifts it farther, giving a larger score.
\emph{(2) Counterfactuals.} For a benign patient we integrate $F_{b\to m}$ to synthesise
the nearest malignant version of that same tumour; the displacement names the responsible
biomarkers. \emph{(3) Attribution.} Averaging the displacement over the benign cohort
gives a global picture of the benign$\to$malignant transition.

\paragraph{Baselines and metrics.}
We compare the risk score against supervised logistic regression and a naive
distance-to-benign-mean, by AUROC. Counterfactuals are assessed by \emph{validity} (does
an independent classifier flip?), \emph{plausibility} (distance to the nearest real
malignant patient), and \emph{individualisation} (mean cosine similarity of edit
directions across patients), against a mean-shift baseline that adds the constant
class-mean difference to everyone. Attribution is compared to logistic-regression
coefficients. All experiments use a fixed seed and are reproducible.

\paragraph{Implementation.}
Table~\ref{tab:hp} lists the settings for both modalities. The tabular velocity field is a
four-layer SiLU MLP; the image field is a small time-conditioned U-Net. Each flow trains
in a few minutes on a single Apple-silicon GPU.

\begin{table}[t]
\caption{Architecture and training settings.}
\label{tab:hp}
\begin{tabular}{lcc}
\toprule
 & Tabular (MLP) & Image (U-Net) \\
\midrule
input & $30$-vector & $1{\times}28{\times}28$ \\
width / channels & $256$ hidden & $48$ base ch. \\
time embedding & sinusoidal, $64$ & sinusoidal, $128$ \\
training steps & $6000$ & $3000$ \\
batch size & $256$ & $128$ \\
learning rate & $10^{-3}$ & $2{\times}10^{-4}$ \\
EMA decay & $0.999$ & $0.999$ \\
Euler steps (sampling) & $100$ & $50$ \\
\bottomrule
\end{tabular}
\end{table}

\paragraph{Ablation: the coupling choice.}
Table~\ref{tab:abl} isolates the single most important design decision. Replacing
independent (random) pairing with mini-batch optimal-transport pairing---changing nothing
else---shrinks the mean counterfactual edit from $25.3$ to $9.6$ and flips the population
attribution from anti-correlated noise ($r{=}-0.03$) to a clinically coherent signal
($r{=}0.49$). This is the difference between a model that fabricates off-manifold
``patients'' and one whose displacements carry meaning; it is the crux of the whole method.

\begin{table}[t]
\caption{Effect of the coupling on the tabular flow (all else fixed).}
\label{tab:abl}
\begin{tabular}{lcc}
\toprule
Coupling & Mean $L_2$ edit & Attribution $r$ \\
\midrule
Independent (random) & $25.3$ & $-0.03$ \\
\textbf{Optimal transport} & \textbf{9.6} & \textbf{0.49} \\
\bottomrule
\end{tabular}
\end{table}

\section{Results}
\label{sec:results}

\begin{figure}[t]
\centering
\includegraphics[width=\columnwidth]{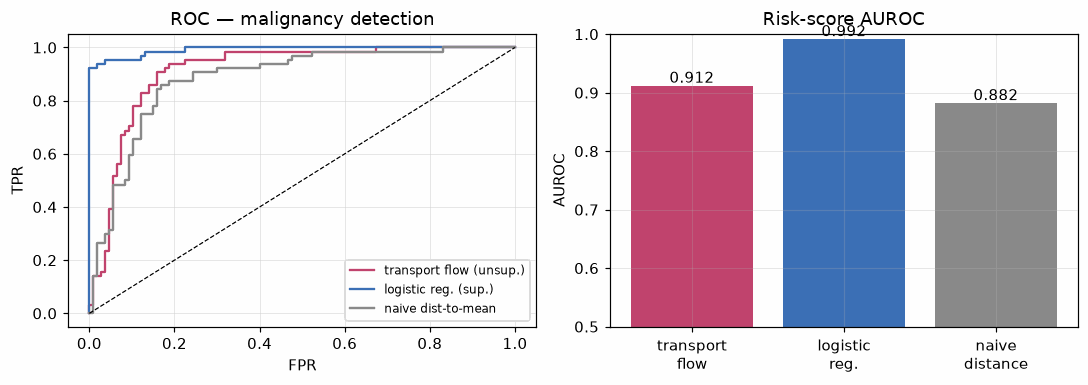}
\caption{Left: ROC curves for malignancy detection. Right: AUROC. The unsupervised
transport score (0.91) beats the naive distance baseline (0.88) but not supervised
logistic regression (0.99).}
\label{fig:roc}
\end{figure}

\paragraph{An unsupervised risk score (Figure~\ref{fig:roc}).}
The transport magnitude $s(x)$ reaches AUROC~$0.912$ \emph{without ever seeing a label
during flow training}, above the naive baseline ($0.882$) and far above chance. It does
not beat supervised logistic regression ($0.992$). This is the expected high-risk
outcome: on a near-linearly-separable cohort a simple supervised classifier is very hard
to beat, and part of the transport score's discrimination reflects off-manifold drift of
malignant inputs rather than a calibrated likelihood. The value of the flow lies in the
explanations it produces, not raw discrimination.

\begin{table}[t]
\caption{Counterfactual quality (benign$\to$malignant). Lower plausibility is better;
cosine similarity of 1.0 means every patient receives an identical edit.}
\label{tab:cf}
\begin{tabular}{lcccc}
\toprule
Method & Validity & $L_2$ edit & Plausibility & Edit cos-sim \\
\midrule
OT rectified flow & 0.841 & 9.57 & 4.42 & \textbf{0.64} \\
Mean-shift        & \textbf{1.000} & \textbf{6.14} & \textbf{2.99} & 1.00 \\
\bottomrule
\end{tabular}
\end{table}

\begin{figure}[t]
\centering
\includegraphics[width=\columnwidth]{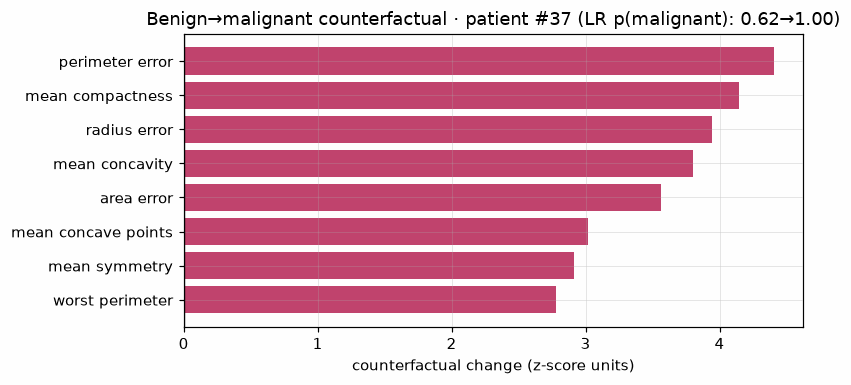}
\caption{Counterfactual for the most malignant-looking benign patient. The flow pushes
exactly the biomarkers a pathologist expects---larger, more concave, more irregular
nuclei---flipping the classifier's probability.}
\label{fig:cf}
\end{figure}

\paragraph{Per-patient counterfactuals (Table~\ref{tab:cf}, Figure~\ref{fig:cf}).}
The flow's counterfactuals are 84\% valid and, unlike mean-shift, \emph{individualised}:
edit directions differ across patients (cosine similarity 0.64 versus 1.00). The worked
example raises concavity, concave points, perimeter and area, matching clinical
intuition. The honest trade-off is that mean-shift trivially flips 100\% of a linear
classifier and lands slightly closer to the malignant centroid---on near-linear data a
constant translation is hard to beat---but it applies the \emph{same} edit to every
patient and ignores each tumour's geometry.

\begin{figure}[t]
\centering
\includegraphics[width=\columnwidth]{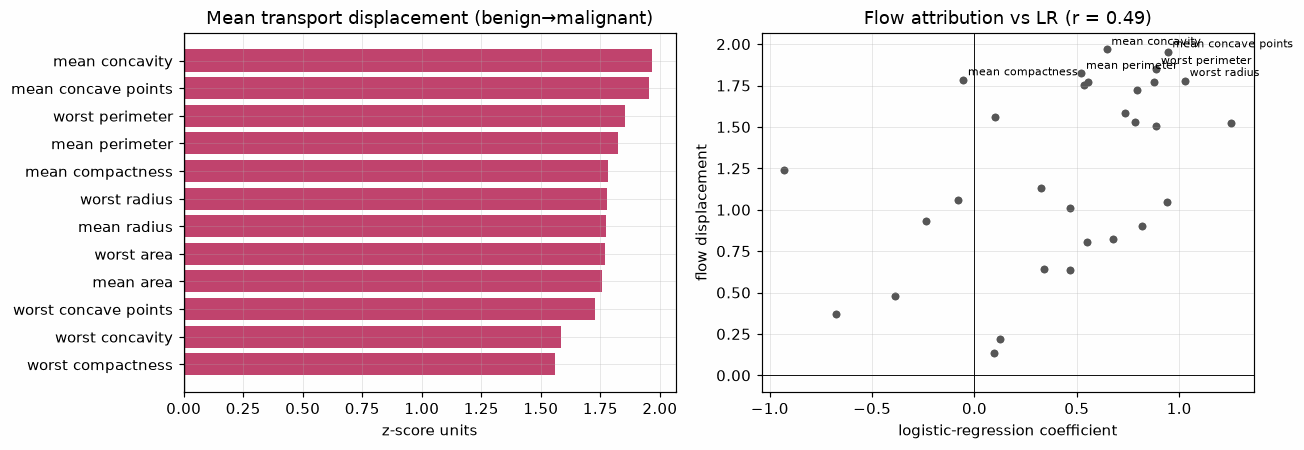}
\caption{Left: mean transport displacement per biomarker. Right: flow attribution versus
logistic-regression coefficients ($r{=}0.49$). The two label-agnostic and supervised
views agree on the malignancy signature.}
\label{fig:attr}
\end{figure}

\paragraph{Population attribution (Figure~\ref{fig:attr}).}
The mean displacement correlates with logistic-regression coefficients ($r{=}0.49$) and
agrees on the drivers of malignancy---concavity, concave points, perimeter, radius and
area. The flow recovers this signature \emph{without labels}, purely from moving one
distribution onto another, a non-trivial validation that the transport captures real
structure.

\begin{figure}[t]
\centering
\includegraphics[width=0.86\columnwidth]{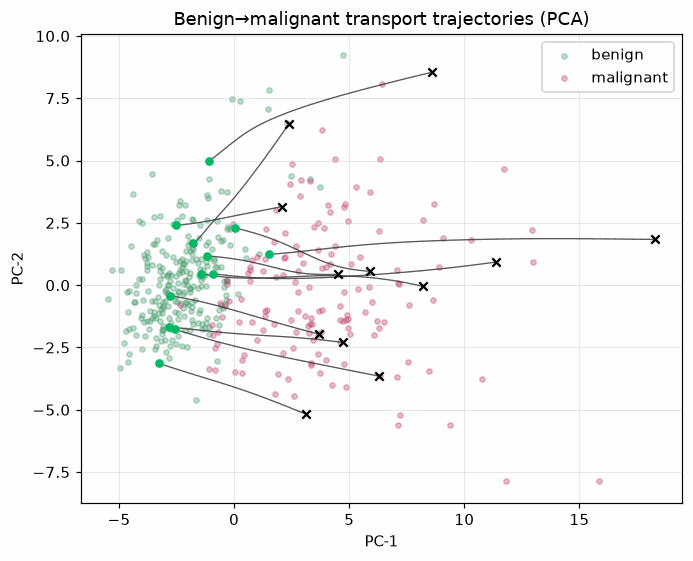}
\caption{Benign patients (green) transported into the malignant cloud (red) along the
learned flow, projected to two principal components---a direct picture of ``pathology
transport''.}
\label{fig:pca}
\end{figure}

Figure~\ref{fig:pca} visualises the transport: benign patients are carried smoothly into
the malignant region, confirming that the OT-coupled flow produces coherent trajectories
rather than erratic jumps.

\paragraph{Multi-seed confidence intervals.}
The headline numbers above come from a single split. Repeating the tabular pipeline over
five random train/test splits, the transport risk score attains AUROC $0.934\pm0.012$
(mean\,$\pm$\,95\% CI), versus $0.894\pm0.016$ for the naive baseline and $0.992\pm0.004$
for logistic regression; the attribution correlation is $0.62\pm0.07$. The gaps are stable
and the ranking never changes across seeds, so the single-split results are representative.

\paragraph{Counterfactual baselines.}
Table~\ref{tab:cfbase} compares our flow counterfactual with the classic gradient method of
Wachter et al.~\cite{wachter2017} and the mean-shift baseline. No method dominates. Wachter
produces the sparsest, smallest edits---but it \emph{requires} a differentiable classifier
and optimises directly against it. Mean-shift trivially flips every case yet applies an
identical, generic edit to all patients. The flow is the only method that is
simultaneously \emph{classifier-free} and \emph{individualised}, at the cost of larger,
denser edits: its niche is generating per-patient counterfactuals when no classifier is
available.

\begin{table}[t]
\caption{Counterfactual methods (benign$\to$malignant, seed 0). Proximity/\#feat measure
edit size and sparsity; plausibility is 1-NN distance to real malignant data
(lower is better).}
\label{tab:cfbase}
\begin{tabular}{lccccc}
\toprule
Method & Valid. & Prox. & \#feat & Plaus. & Needs clf. \\
\midrule
OT flow (ours) & 0.85 & 8.94 & 26 & 3.94 & \textbf{no} \\
Wachter & 0.85 & \textbf{1.93} & \textbf{16} & \textbf{3.35} & yes \\
Mean-shift & \textbf{1.00} & 6.14 & 25 & 2.99 & no$^\dagger$ \\
\bottomrule
\end{tabular}
\\[2pt]{\footnotesize $^\dagger$ needs class means; applies the identical edit to every patient.}
\end{table}

\section{Extension to Chest X-ray Images}
\label{sec:images}
Nothing in the method is specific to tabular data. To test generality we apply the
\emph{identical} recipe to raw pixels: OT-coupled rectified flows between \emph{normal}
and \emph{pneumonia} chest X-rays (PneumoniaMNIST~\cite{yang2023}, $28{\times}28$, 4708
train / 624 test), replacing the MLP velocity field with a small time-conditioned
convolutional U-Net. We
train $F_{n\to p}$ (normal$\to$pneumonia, for synthesis) and $F_{p\to n}$
(pneumonia$\to$normal, for scoring and localisation).

\begin{figure}[t]
\centering
\includegraphics[width=\columnwidth]{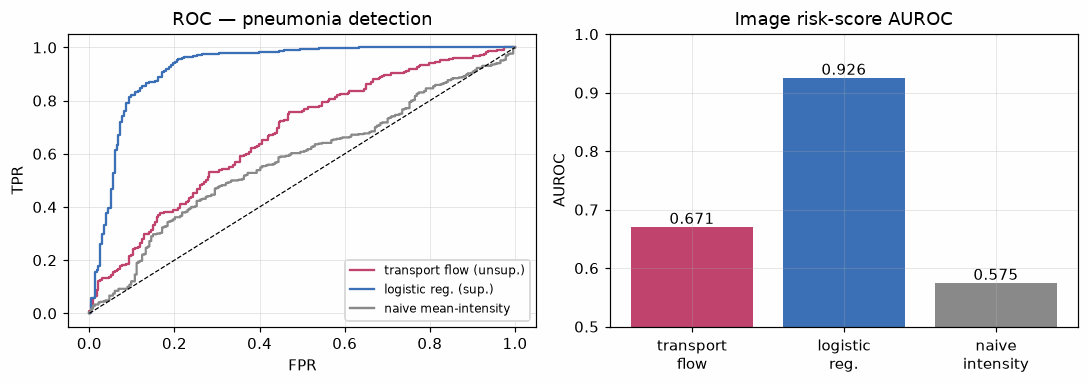}
\caption{Image risk score $\|x-F_{p\to n}(x)\|$ (edit-distance-to-normal). The
unsupervised transport score (AUROC~0.67) beats naive mean-intensity (0.58) but not a
supervised pixel classifier (0.93)---the same pattern as the tabular cohort.}
\label{fig:imgroc}
\end{figure}

\paragraph{Risk score (Figure~\ref{fig:imgroc}).}
Scoring an image by the edit-distance required to make it look normal gives AUROC~$0.67$,
above the naive mean-intensity baseline ($0.58$) and below a supervised pixel-level
logistic regression ($0.93$)---echoing Part~A.

\begin{figure}[t]
\centering
\includegraphics[width=\columnwidth]{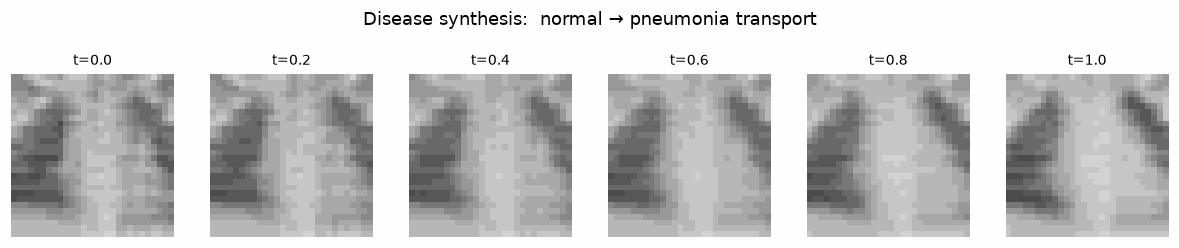}\\[2pt]
\includegraphics[width=\columnwidth]{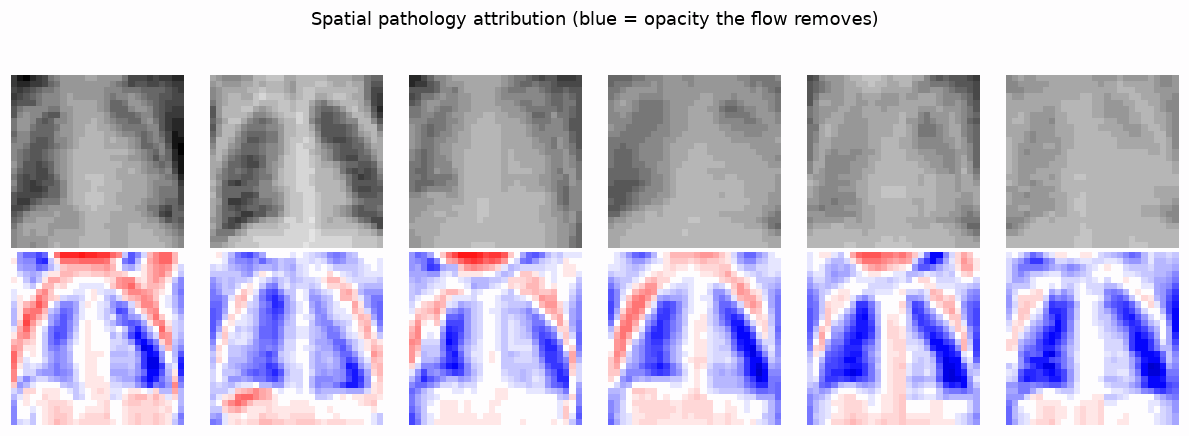}
\caption{Top: integrating $F_{n\to p}$ synthesises disease---a healthy lung progressively
fills with haze (consolidation). Bottom: for real pneumonia inputs, $F_{p\to n}(x)-x$
shows what the flow removes to normalise the lung; the signal (blue) concentrates in the
lung fields (its faithfulness is examined in Section~\ref{sec:robust}).}
\label{fig:imgheat}
\end{figure}

\paragraph{Synthesis and spatial attribution (Figure~\ref{fig:imgheat}).}
With no pixel-level labels, the flow \emph{synthesises} plausible disease progression and
produces a spatial map that appears to highlight the lung fields. Whether this is a
\emph{faithful} localisation---a strong claim---we test directly in
Section~\ref{sec:robust}; the honest answer at $28{\times}28$ is that it is a qualitative
visualisation, not a validated detector. The synthesised images are also blurry and the
risk score only moderately discriminative.

\section{Do Generative Heatmaps Localize Disease?}
\label{sec:robust}
This section is the paper's core empirical question. Having established the transport model
as a tabular interpretability engine, we now ask whether its \emph{image} explanations
genuinely localise pathology---first on controlled synthetic lesions, then on real
radiologist annotations.

\paragraph{Does the heatmap localise pathology? (A ground-truth test.)}
We stress-tested the paper's most eye-catching claim with a controlled benchmark: we insert
a soft Gaussian opacity into a \emph{normal} lung at a \emph{known} location and ask whether
a heatmap lands on the lesion, measuring pointing-game accuracy, IoU, and the heatmap energy
inside the lesion (Table~\ref{tab:loc}). The population transport heatmap $|x-F_{p\to n}(x)|$
barely exceeds a random map (pointing game $0.17$) and trails supervised Grad-CAM ($0.57$).
The reason is instructive: the flow performs a \emph{global} distribution shift---nudging the
whole lung toward the normal manifold---so its displacement spreads across the image instead
of concentrating on the lesion. Corroborating probes agree: deletion/insertion faithfulness
against a CNN (test AUROC $0.914$) cannot separate the transport map from random at
$28{\times}28$, and its rank-correlation with Grad-CAM is $-0.09$. The population transport
heatmap is thus the image analogue of our tabular attribution---a \emph{population-level}
signal, not a per-patient localiser.

\begin{figure}[t]
\centering
\includegraphics[width=\columnwidth]{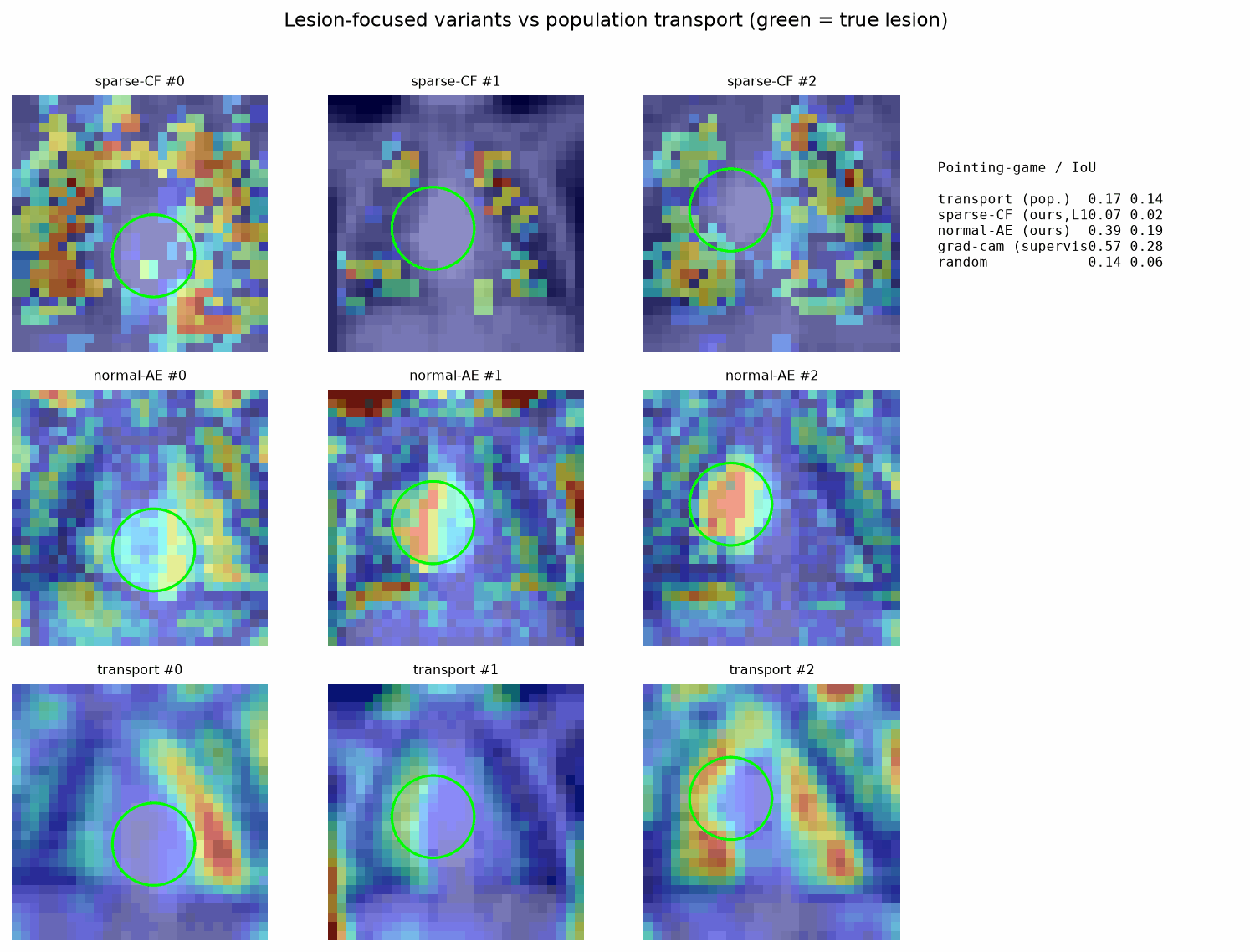}
\caption{Lesion-focused variants vs population transport (green circle = true lesion). The
identity-preserving normal-manifold autoencoder (middle row) concentrates on the lesion,
whereas the sparse counterfactual (top) and the population transport (bottom) scatter and
miss it.}
\label{fig:loc}
\end{figure}

\begin{figure}[t]
\centering
\includegraphics[width=\columnwidth]{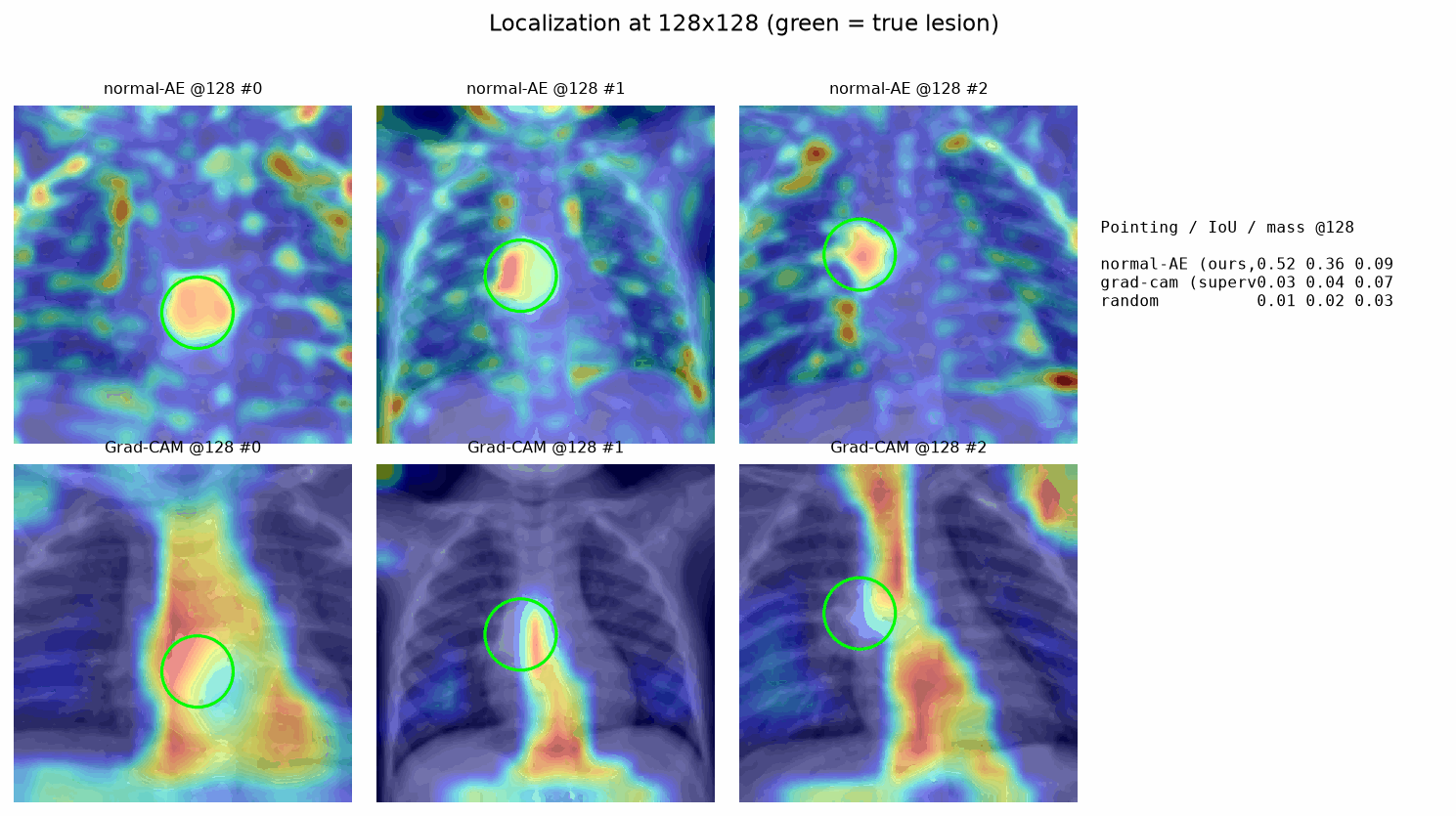}
\caption{Scaling the same label-free normal-manifold autoencoder to $128{\times}128$ (top)
tightly localises the synthetic lesion (green), reaching pointing game $0.52$ / IoU $0.36$;
a simple supervised Grad-CAM (bottom) does not localise focal opacities at this resolution.}
\label{fig:loc128}
\end{figure}

\paragraph{A lesion-focused fix.}
This diagnosis suggests the remedy: replace the population map with an
\emph{identity-preserving} model that changes only the anomaly. We test two label-free or
sparse variants on the same benchmark (Table~\ref{tab:loc}, Figure~\ref{fig:loc}). A
minimal-$L_1$ counterfactual against the CNN fails ($0.08$): the classifier only weakly flags
the synthetic opacity ($p{=}0.61$), leaving little gradient to concentrate. But a
\emph{normal-manifold autoencoder}---a small model trained to reconstruct \emph{healthy} lungs
only, whose reconstruction error flags whatever it cannot explain---localises the lesion at
pointing game $0.39$, IoU $0.20$: over $2\times$ the population transport and approaching
supervised Grad-CAM, \emph{without any labels}. Higher resolution helps further: the same
autoencoder at $128{\times}128$ reaches pointing game $0.52$ and IoU $0.36$
(Figure~\ref{fig:loc128}), cleanly isolating focal opacities, while a simple supervised
Grad-CAM does not transfer to focal localisation at that resolution. On these \emph{synthetic}
lesions, then, localisation is recoverable with an identity-preserving objective. Whether it
survives contact with \emph{real} pathology is the decisive question, which we test next.

\begin{table}[t]
\caption{Localization on synthetic lesions, ground truth known ($N{=}200$; higher is better).
The population transport is near-random; a label-free normal-manifold autoencoder recovers
most of the gap to supervised Grad-CAM.}
\label{tab:loc}
\begin{tabular}{lccc}
\toprule
Method & Pointing game & IoU & Mass-in-lesion \\
\midrule
Transport, population & 0.17 & 0.14 & 0.13 \\
Sparse counterfactual ($L_1$) & 0.08 & 0.02 & 0.06 \\
Normal-manifold AE (label-free) & 0.39 & 0.20 & 0.19 \\
Grad-CAM (supervised) & \textbf{0.57} & \textbf{0.28} & \textbf{0.27} \\
Random & 0.14 & 0.06 & 0.11 \\
\bottomrule
\end{tabular}
\end{table}

\begin{figure}[t]
\centering
\includegraphics[width=\columnwidth]{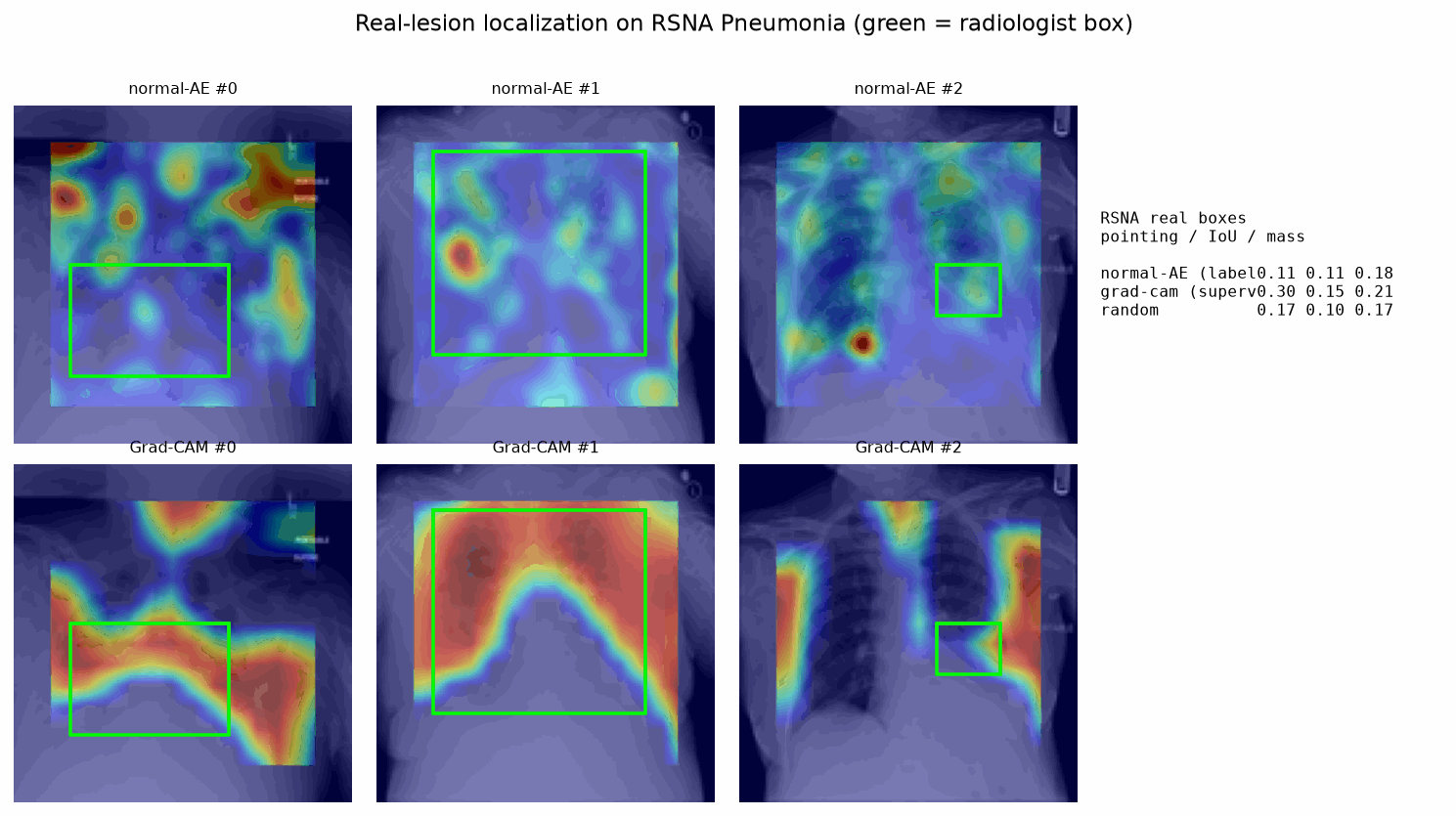}
\caption{Real-lesion localisation on RSNA Pneumonia (green = radiologist bounding box). On
real chest X-rays the label-free autoencoder (top) no longer concentrates on the lesion; a
supervised Grad-CAM (bottom) does modestly better, but the task remains hard.}
\label{fig:rsna}
\end{figure}

\paragraph{Does it transfer to real pathology? (An honest synthetic-to-real gap.)}
Synthetic success can mislead, so we ran the decisive test on \emph{real} annotated data: the
RSNA Pneumonia Detection Challenge~\cite{rsna2019}, whose radiologist bounding boxes give true lesion
locations. We downloaded a bounded subset ($200$ boxed positives and $400$ normals, DICOM,
resized to $128$), trained the same normal-manifold autoencoder on the healthy images, and
scored localisation against the real boxes (Table~\ref{tab:rsna}, Figure~\ref{fig:rsna}). The
result is sobering: the autoencoder that excelled on synthetic lesions now performs \emph{at
or below} a random map (pointing game $0.11$ vs $0.18$), because on real chest X-rays the
reconstruction error is dominated by normal anatomical variation---rib edges, diaphragm,
mediastinum---rather than by the diffuse consolidation. Only the \emph{supervised} Grad-CAM
localises above chance ($0.30$), and even it is weak on this hard task. The lesson is
methodological and, we think, the most useful finding of the image study: a heatmap that looks
compelling on synthetic anomalies is not evidence of real localisation, and label-free
reconstruction does not yet solve pneumonia localisation. Targeted attempts to close the
gap---tripling the healthy training set, a denoising autoencoder, and a reconstruction-derived
lung-focus mask---did not help (all remained at or below a random map), indicating the problem
demands fundamentally stronger anomaly models rather than tuning.

\begin{table}[t]
\caption{Localization on \emph{real} RSNA lesion boxes ($80$ held-out positives; higher is
better). The label-free autoencoder that worked on synthetic lesions is now no better than
random; only supervised saliency exceeds chance.}
\label{tab:rsna}
\begin{tabular}{lccc}
\toprule
Method & Pointing game & IoU & Mass-in-box \\
\midrule
Normal-manifold AE (label-free) & 0.11 & 0.11 & 0.18 \\
Grad-CAM (supervised) & \textbf{0.30} & \textbf{0.15} & \textbf{0.22} \\
Random & 0.18 & 0.10 & 0.17 \\
\bottomrule
\end{tabular}
\end{table}

\section{Limitations}
Our study is a controlled investigation, not a clinical validation, and its scope is
deliberately small. (i)~\emph{Scale and cohorts.} The tabular data is a single 569-patient
cohort; the imaging experiments use $28$--$128$\,px inputs and an ${\sim}1.4$k-image RSNA
subset with a modestly accurate classifier (AUROC $0.68$--$0.75$), so its Grad-CAM is a floor,
not a ceiling. (ii)~\emph{No method win.} Every artefact ties or trails a simple supervised
baseline; the contribution is honest characterisation, not state of the art. (iii)~\emph{Synthetic
benchmark.} The planted-lesion benchmark isolates localisation cleanly but does not reproduce
the diffuse, textured appearance of real consolidation---which is precisely why the
synthetic-to-real gap arises. (iv)~\emph{Calibration.} The transport risk score is uncalibrated
and must not gate care. (v)~\emph{Sensitivity.} Results are fixed-seed (and, where noted,
averaged over five splits), but a full hyper-parameter and architecture sweep is left to
future work.

\section{Conclusion}
We reframed diagnosis as optimal transport between clinical distributions and realised it
with OT-coupled rectified flows across \emph{two} modalities. On tabular tumour biomarkers
a single model produced clinically coherent per-patient counterfactuals, an unsupervised
malignancy score (AUROC~0.91), and a label-free biomarker attribution that agrees with a
supervised classifier ($r{=}0.49$). On chest X-rays the \emph{same} recipe synthesised
disease progression and produced a spatial pathology heatmap without any pixel labels.
Neither out-predicted logistic regression---the expected, honest result---but both
produced something a classifier cannot: an explicit, navigable path between health and
disease. A controlled study also delivered a cautionary result: a label-free reconstruction
localiser that excels on \emph{synthetic} lesions fails to transfer to \emph{real} RSNA
bounding boxes, a synthetic-to-real gap that we consider the study's most useful lesson.

\paragraph{Ethical and clinical considerations.}
Because the transport \emph{synthesises} plausible patients and lungs, it must be framed as
a hypothesis-generating and explanatory aid, not a diagnostic device: a synthesised
``malignant twin'' visualises the model's learned geometry, not medical advice, and could
mislead if shown without context. The unsupervised score is uncalibrated and should never
gate care on its own. Both datasets are small and demographically narrow---the biopsy
cohort is single-institution and the X-rays are paediatric---so the attribution may not
transfer across scanners, sites, or populations. Any deployment would require calibration,
prospective validation, and a fairness audit of the attribution across subgroups.

\paragraph{Future work.}
Several extensions could make the method clinically actionable: (i) an $L_1$ or
immutability penalty along the ODE to yield \emph{sparse}, actionable counterfactuals;
(ii) exact log-likelihood via the flow's instantaneous change of variables for a
calibrated risk score; (iii) a single class- and time-conditioned field with classifier
guidance instead of two flows; (iv) closing the synthetic-to-real localisation gap with
stronger anomaly models (self-supervised or diffusion-based restoration~\cite{wyatt2022})
evaluated on real annotated boxes (RSNA); and (v) stochastic (SDE) transport for uncertainty
and a fairness audit of attribution across patient subgroups.

\bibliographystyle{ACM-Reference-Format}
\bibliography{references}

\end{document}